\documentclass[10pt,twocolumn,letterpaper]{article}

\usepackage{cvpr}
\usepackage{times}
\usepackage{epsfig}
\usepackage{graphicx}
\usepackage{amsmath}
\usepackage{amssymb}
\usepackage{caption}
\usepackage{subcaption}

\usepackage[utf8]{inputenc}
\usepackage[english]{babel}
\usepackage{algorithm}
\usepackage{algpseudocode}
\usepackage{multirow}

\newcommand{\bfsection}[1]{\vspace*{0.1cm}\noindent\textbf{#1.}}
\newcommand{\red}[1]{\textcolor{red}{#1}}

\usepackage{color}
\definecolor{citecolor}{RGB}{34,139,34}
\usepackage[pagebackref=true,breaklinks=true,letterpaper=true,colorlinks,bookmarks=false]{hyperref}
\hypersetup{
     colorlinks = true,
     linkcolor = red,
     anchorcolor = black,
     citecolor = citecolor,
     filecolor = black,
     urlcolor = black
     }

\cvprfinalcopy 

\def\cvprPaperID{19} 

\newcommand*{\email}[1]{{\small\texttt{#1}}}
\begin{document}

\title{A Topological Nomenclature for 3D Shape Analysis in Connectomics}

\author{
Abhimanyu Talwar\qquad 
Zudi Lin\qquad
Donglai Wei$^{*}$\qquad
Yuesong Wu\qquad
Bowen Zheng\\
Jinglin Zhao\qquad
Won-Don Jang\qquad
Xueying Wang\qquad
Jeff Lichtman\qquad
Hanspeter Pfister\vspace{0.1in}\\
Harvard University\\
\email{donglai@seas.harvard.edu}
}

\twocolumn[{
\renewcommand\twocolumn[1][]{#1}
\maketitle
\thispagestyle{empty}
    \centering
        \includegraphics[width=0.9\textwidth]{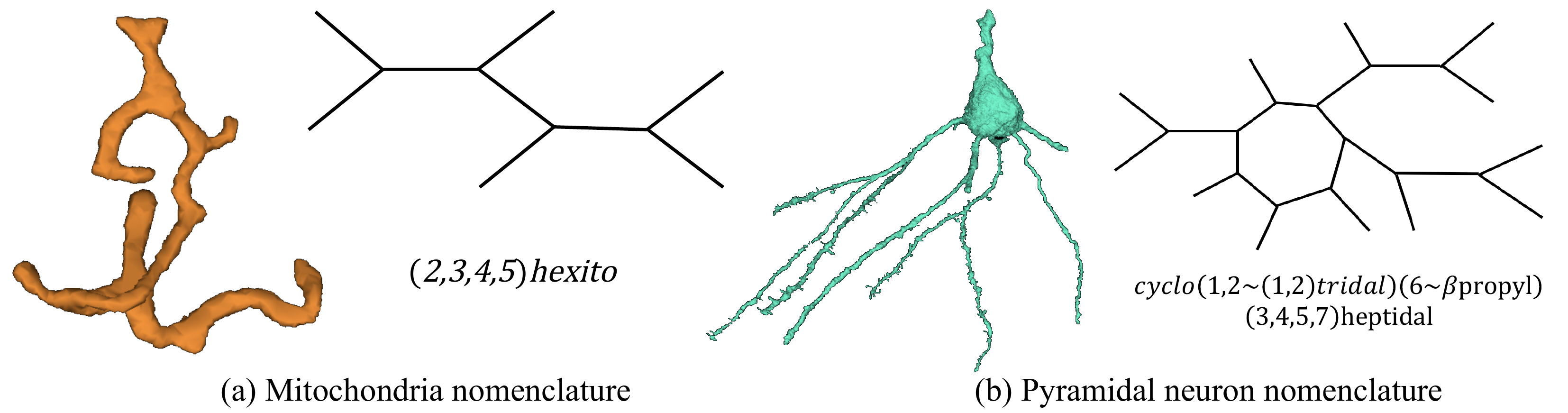} 
    \captionof{figure}[]{Illustration of topological nomenclature for mitochondria and pyramidal neurons in EM connectomics.
    Given the input 3D segmentation, our proposed algorithm generates reduced graph representation to capture essential morphological structure and nomenclature for concise scientific communication.}
    \label{fig:intro}
    \vspace{1.5em}
}]

\begin{abstract}
One of the essential tasks in connectomics is the morphology analysis of neurons and organelles like mitochondria to shed light on their biological properties. However, these biological objects often have tangled parts or complex branching patterns, which make it hard to abstract, categorize, and manipulate their morphology.  In this paper, we develop a novel topological nomenclature system to name these objects like the appellation for chemical compounds to promote neuroscience analysis based on their skeletal structures. We first convert the volumetric representation into the topology-preserving reduced graph to untangle the objects. Next, we develop nomenclature rules for pyramidal neurons and mitochondria from the reduced graph and finally learn the feature embedding for shape manipulation. In ablation studies, we quantitatively show that graphs generated by our proposed method align with the perception of experts. On 3D shape retrieval and decomposition tasks, we qualitatively demonstrate that the encoded topological nomenclature features achieve better results than state-of-the-art shape descriptors. To advance neuroscience, we will release a 3D segmentation dataset of mitochondria and pyramidal neurons reconstructed from a 100$\mu$m cube electron microscopy volume with their reduced graph and topological nomenclature annotations. Code is publicly available at \url{https://github.com/donglaiw/ibexHelper}.
\end{abstract}

\section{Introduction}
Recent advancements in large-scale electron microscopy (EM) allows the generation of petabytes of serial images of brain tissue at nanometer resolution~\cite{kasthuri2015saturated,zheng2018complete}. Machine learning methods have made automated 3D reconstruction possible for individual neurons~\cite{januszewski2018high} and intracellular organelles such as mitochondria~\cite{cheng2017volume}.  Intriguingly, 3D shapes of these objects resolved at the nano-scale are far more complicated than the classic depiction in the neuroscience textbooks. Thus, novel morphology analysis tools are required to advance our understanding of the basic properties of neuronal compartments (Fig.~\ref{fig:intro}).

Nevertheless, there are three main challenges. First, the branches and loops of a non-convex object can tangle together in the 3D meshes, which make it difficult for an intuitive perception of the underlying topology. Second, there lacks an intuitive and concise way to convey the shape information of neurons and organelles in the neuroscience community. Third, the goal of traditional descriptors of 3D meshes is to compare objects with similar scales, which is not suitable for the application on neurons and organelles that have a wide range of spatial dimensions.

\begin{figure*}[t]
     \centering
     \includegraphics[width=1.0\textwidth]{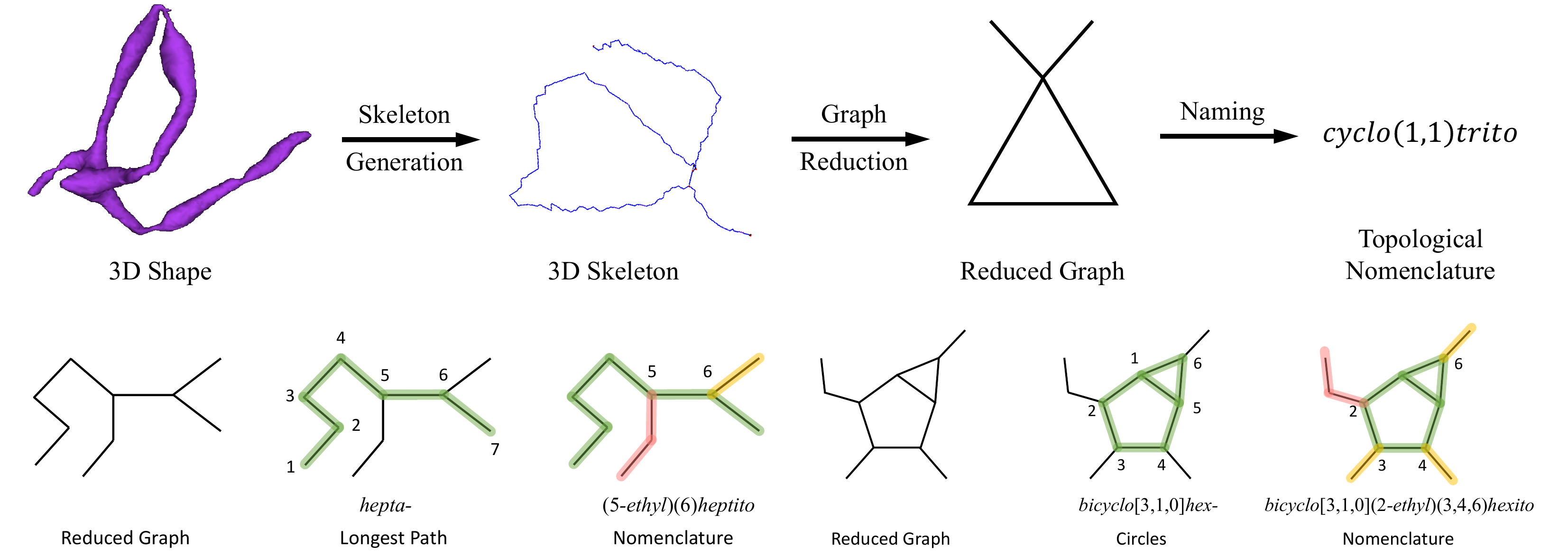}
     \caption{Overview of our topological nomenclature framework (first row) and nomenclature rules (second row). Best view in color.}
     \label{fig:method}
\end{figure*}

To tackle these challenges, we propose a topological nomenclature system to abstract, categorize, and manipulate the 3D meshes of neurons and organelles (Fig.~\ref{fig:intro}).  We first {\em skeletonize} them into vertexes and edges to untangle the objects. We further prune them into a concise reduced-graph while preserving the topological properties. To systematically name those objects, we propose a nomenclature system borrowing ideas from nomenclature for organic compounds. The primary aim of nomenclature in chemistry is to ensure that every name refers to a specific compound without ambiguity.  The naming systems, including InChI~\cite{heller2013inchi} and SMILES, display more structural details but is more cumbersome for scientific communication. Therefore in this work, we follow the IUPAC rule~\cite{panico1993guide} to generate the graph name that is more human-readable. To apply the nomenclature system to shape analysis, we use the deep learning model for self-supervised learning on graphs~\cite{kipf2016variational}. In comparison with traditional shape descriptors like heat kernel signature (HKS)~\cite{5457682}, one key characteristic of our approach is that our graph representations are more intuitive to understand than the HKS and allow for a simple shape decomposition into primitives. Sundar \emph{et al.}~\cite{Sundar:2003:SBS:829510.830308} explore the idea of the skeleton-based 3D shape matching. They introduce the concept of a topological signature vector - a low dimensional representation of a graph that can be a measure of similarity. One difference from our approach is that their scheme generates an acyclic skeletal graph that does not capture cycles or multiple paths between two vertices.

To summarize, we present three main contributions in this paper. First, we propose a shape abstraction method that converts 3D meshes into 2D graphs using a novel combination of skeletonization and graph reduction to improve the morphological perception. Second, by using the nomenclature system, we not only make it more interpretable by neuroscientists but also further compress the information needed for graph reconstruction. Third, we implement an unsupervised model to embed these graphs into vector space for 3D shape retrieval and decomposition.

\section{Related Work}
\bfsection{Chemical and Biological Nomenclature}
Organic molecules, which contain carbon as the backbone, exhibit a variety of structures. Therefore a concise and intuitive nomenclature system is crucial for informative scientific communication. The primary aim of nomenclature in chemistry and biology is to ensure that every name refers to a specific compound without ambiguity. The secondary aim is that the name can (to some extent) reflects a substance's structure. There are two main streams of chemical nomenclature. The first stream that follows the IUPAC rule~\cite{panico1993guide} is relatively simple and more human-readable. Other systems, including InChI~\cite{heller2013inchi} and SMILES, display more structural details in the name but is more cumbersome for scientific communication. For large organic molecules like proteins that form a more complex spatial arrangement, Flower~\cite{flower1998topological} proposes to abstract the modules into graphs and name them based on topology. Our work extends the nomenclature to neurons and mitochondria morphologies in EM connectomics that do not have well defined functional groups (e.g., benzene) and shape primitives (e.g., $\alpha-$ and $\beta-$helix of proteins) as biochemical compounds.

\begin{figure*}[t]
     \centering
     \includegraphics[width=0.8\textwidth]{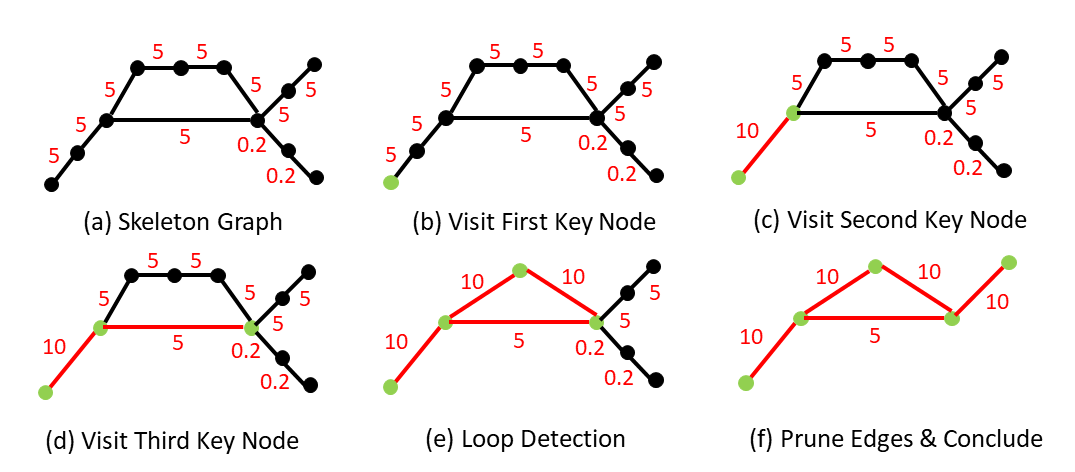}
     \caption{Illustration of graph reduction. We modify the breadth-first search (BFS) traversal to consider only key nodes in the graph. Unlike BFS which results in a tree, our traversal preserves loops within the graph to maintain the cyclic topology of some mitochondria and pyramidal neurons that form loops. Please see more details in Algorithm~\ref{algo:traversal}.}
     \label{fig:bfs_example}
\end{figure*}

\bfsection{3D Shape Analysis in Connectomics}
Recent advances in EM imaging have enabled connectomics research at nano-scale resolution. For example, the 3D instance masks of thousands of neurons and millions of intracellular organelles like mitochondria are available for analysis. However, earlier studies only conduct basic statistical analysis, including the length, volume, and surface-to-volume ratio~\cite{kasthuri2015saturated}. Recent Kanari \emph{et al.}~\cite{kanari2018topological} propose a topological morphology descriptor of neuronal structures based on distance transform to analyze the shape of pyramidal neurons. However, such a method is not intuitive for human perception as the reduced distance maps can hardly reflect the topology of the original instances.

\bfsection{3D Shape Descriptors}
Matching and retrieval of 3D shapes are mature disciplines, and various successful schemes are out there. For example, Ovsjanikov \emph{et al.}~\cite{5457682} to create an isometry invariant shape retrieval system by adopting a heat kernel signature (HKS) based deformation invariant shape descriptor~\cite{article}. Sundar \emph{et al.}~\cite{Sundar:2003:SBS:829510.830308} develop the skeleton-based 3D shape matching algorithm and define a topological signature vector, which is a low dimensional representation of a graph.  These two methods are different from our approach in two ways. First, our graph representations are more understandable and intuitive compared to HKS while allowing a simple shape decomposition into primitives.  Unlike the existing skeleton-based 3D shape matching scheme that generates an acyclic skeletal graph, our algorithm is capable of capturing cycles or multiple paths between two vertices.


\section{Method}
In this section, we give a formal definition of our shape abstraction and nomenclature system (Fig.~\ref{fig:method}). Given an input 3D mesh, we first transform it into a reduced graph that preserves its topological information with a novel skeletonization algorithm. We then determine its name in the nomenclature system based on its category (e.g., mitochondria) and graph structure. We also show how to compute the object feature in the nomenclature embedding space for the following manipulation. 
\subsection{Topology-Aware Reduced Graph Generation}\label{subsec:graphgen}
Starting from the volumetric representation of a 3D mesh, we convert it into a reduced graph which preserves its topological structure, like the molecular graph~\cite{mcnaught1997compendium}, using the following three steps:

\bfsection{Graph Initialization} 
We use an off-the-shelf skeletonization algorithm proposed in K{\'a}lm{\'a}n~\cite{10.1007/978-3-319-14364-4_39} to extract a 3D skeleton from the voxel representation. We can view the extracted 3D skeleton as a weighted undirected graph $G=(V, E, W)$ where $V \subset \mathbb{Z}^{*3}$ is the set of coordinates of skeleton nodes in the 3D voxel grid, $E \subset V \times V$ is the set of edges, and $W \subset \mathbb{R}^+$ is the set of edge weights.

\bfsection{Graph Reduction}
Based on the degree of incident edges, we can divide the skeleton vertices $V$ into junctions $J = \{n \in V : degree(n) > 2\}$ and endpoints $E = \{n \in V : degree(n) = 1\}$. We aim to reduce the skeleton graph $G$ to a graph $G_S$ whose set of vertices is $J \cup E$ (referred to as the \textit{key nodes}), and which preserves topological features of $G$ such as paths and distances (along with the 3D skeleton) between any pair of key nodes. Further, we also require $G_S$ to preserve any cycles present in the skeleton graph $G$ and to preserve multiple paths between any two key nodes.

\begin{figure*}[t]
     \centering
     \includegraphics[width=0.8\textwidth]{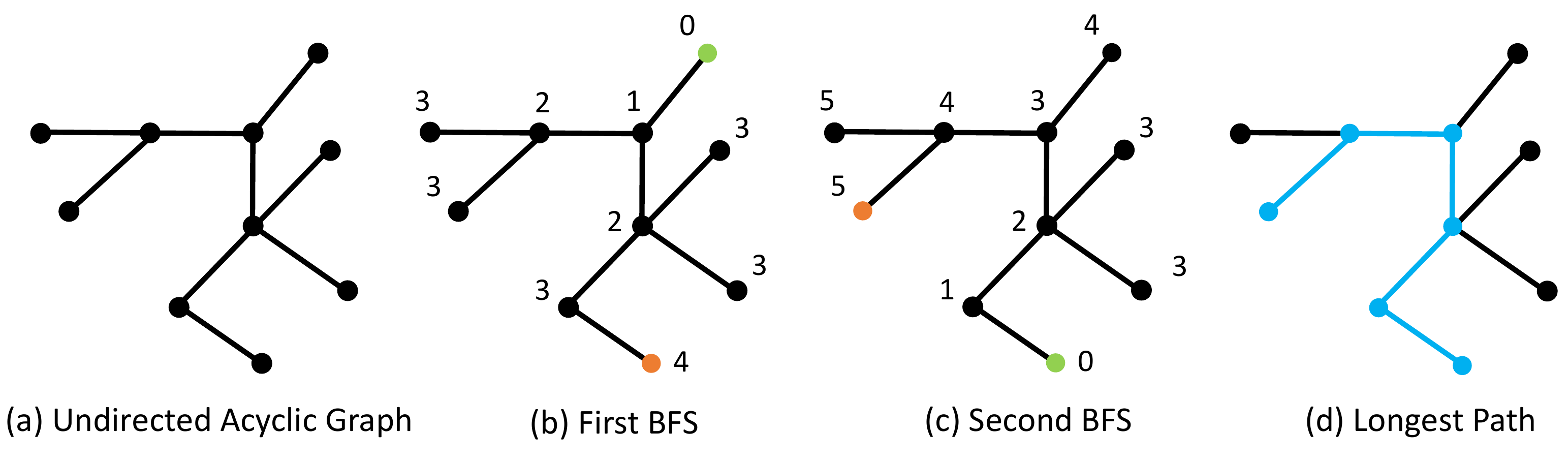}
     \caption{Illustration of extracting the longest path for the nomenclature assignment of undirected acyclic graphs (UAG) (Sec.~\ref{sec:nomenclature}). Although the longest-path problem in an arbitrary graph is $NP$-complete, the problem can be efficiently solved by running the breadth-first search (BFS) algorithm twice for UAGs. The first pass starts from an arbitrary leaf point (b), while the second pass starts from the farthest point found in the first round (c).}
     \label{fig:longest_path}
\end{figure*}

\begin{table*}[t]
\centering
\caption{Prefixes and suffixes in our nomenclature system. We use Greek numeral prefixes to indicate the number of nodes on the longest path in an acyclic graph or on the rings, and use suffixes to categorize different type of instances.}
\label{tab:prefix}
\begin{tabular}{llllllllll|cc}
\hline
\multicolumn{10}{c|}{Prefixes (number of key points)} & \multicolumn{2}{c}{Suffixes (type)}\\
\hline
1 & 2 & 3 & 4 & 5 & 6 & 7 & 8 & 9 & 10 & Mitochondira & Pyramidal Cell\\
\hline
{\em mono}- & {\em di}- & {\em tri}- & {\em tetra}- & {\em penta}- & {\em hexa}- & {\em hepta}- & {\em octo}-  & {\em ennea}- & {\em deca}- & -{\em ito} & -{\em idal}\\
\hline
\end{tabular}
\end{table*}
 
For graph reduction, we modify the \textit{Breadth-First-Search} traversal algorithm, as outlined in Algorithm~\ref{algo:traversal}. At each step of the traversal, we only enqueue key nodes to our traversal queue. We initialize the queue with any key node, and while visiting a key node $v$, we only enqueue (1) any key nodes which are adjacent to $v$, and (2) any other key nodes which are connected to $v$ by a path in $G$ comprising only of non-key nodes (referred to as a \textit{simple path}). 
We define the ``thickness" of an edge as the average of distance transforms of its two vertices, and the ``thickness" of a path can be calculated as the mean of the thickness of each edge on the path weighted by edge lengths. During the traversal, we keep track of two metrics for every pair of key nodes connected by a simple path: (1) sum of lengths of all edges in $G$ on that simple path, and (2) mean thickness of that simple path. An illustration of the graph reduction scheme is shown in Fig.~\ref{fig:bfs_example}.

\bfsection{Graph Post-procession}
In the biological systems, larger structures (\eg, large synapses with more vesicles and higher post-synaptic densities) usually contribute more to the overall functionality. Therefore to further simplify the graph without the distraction caused by numerous small structures, we can remove edges and cycles whose path length is small relative to the total length of edges in $G_S$. Thus, we contract all edges with a length lower than a threshold value of $\tau$. With bigger $\tau$, we obtain a coarser-level representation of the graph. In the experiments (Sec.~\ref{sec:exp}), we perform ablation studies to demonstrate how the threshold $\tau$ affect the quality of the reduced graphs in terms of the agreement with human perception.

\subsection{Topological Nomenclature Rules}\label{sec:nomenclature}
Our nomenclature system for EM connectomics is modified upon the IUPAC nomenclature of organic chemistry, which is not only invariant to the deformation and the graph indexing order but also easily convertible back to the graph representation. We add suffixes \textbf{\emph{-ito}} and \textbf{\emph{-idal}} to mitochondria and pyramidal neurons, respectively.

\bfsection{Acyclic Graph} An acyclic graph is a graph having no cycles. A reduce graph can be entirely a tree structure or contain tree branches. The nomenclature rule for a tree is first to count the longest chain of vertexes, and assign a prefix based on the number of vertexes. For example, the longest chain with $n=5$ vertexes has a prefix \textbf{\emph{penta}} (Table~\ref{tab:prefix}). Find the longest path in a general graph has been shown to be a $NP$-complete problem~\cite{cormen2009introduction}, but find the longest path in an undirected tree graph can be solved efficiently by running the breadth-first search (BFS) algorithm twice (see Fig.~\ref{fig:longest_path} for detail). Therefore such a rule makes sure that for acyclic graphs, not only computer programs but also human users can efficiently drive the corresponding topological nomenclature. Then every vertex on the longest path is assigned a location number from $1$ to $n$. For a branch, we use the location index as prefix and name the branch recursively based on the rules. For simplicity, we combine the prefix of branches with the same structure and omit the description for branch topology if a branch only contains one node. An example of the naming of the acyclic structure is shown in Fig.~\ref{fig:method}.

\begin{algorithm*}
\caption{Skeleton Graph Traversal}\label{algo:traversal}
    \begin{algorithmic}[1]
        \Procedure{ReduceGraph}{}
        \State{$Q \gets \text{Any }v \in \left(J \cup P\right)$}
        \State{$V_S, E_S, W_S \gets \{\}, \{\}, \{\}$}
        \State{$G_S \gets (V_S, E_S, W_S)$}
        \State{$visited[v] \gets \text{False } \forall v \in V$}
        \While{not Q.isEmpty()}
            \State{$src \gets Q.dequeue()$}
            \State{$visited[src] \gets \text{True}$}
            \State{$adj\_list \gets \text{KEY\_NEIGHBORS}\left(G, src\right)$}
            \For{$trg \in adj\_list$}
                \If{not visited[trg]}
                    \State{$weight \gets \text{PATH\_LEN}(src, trg)$}
                    \If{$(src, trg) \notin E_S$}
                        \State{$G_S \gets \text{ADD\_EDGE}(G_S, (src, trg), weight)$}
                    \Else
                        \Comment{True if multiple paths exist between $src$ and $trg$}
                        \State{$mid \gets \text{NEW\_KEY\_NODE()}$}
                        \State{$G_S \gets \text{ADD\_EDGE}(G_S, (src, mid), weight/2)$}
                        \State{$G_S \gets \text{ADD\_EDGE}(G_S, (mid, trg), weight/2)$}
                    \EndIf
                    \If{$trg \notin Q$}
                        \State{$Q \gets Q.enqueue(trg)$}
                    \EndIf
                \EndIf
                \If{src == trg}
                    \Comment{True if $G$ has a cycle.}
                    \State{$mid1 \gets \text{NEW\_KEY\_NODE()}$}
                    \State{$mid2 \gets \text{NEW\_KEY\_NODE()}$}
                    \State{$G_S \gets \text{ADD\_EDGE}(G_S, (src, mid1), weight/3)$}
                    \State{$G_S \gets \text{ADD\_EDGE}(G_S, (mid1, mid2), weight/3)$}
                    \State{$G_S \gets \text{ADD\_EDGE}(G_S, (mid2, trg), weight/3)$}
                \EndIf
            \EndFor{}
        \EndWhile
        \EndProcedure{}
    \end{algorithmic}
\end{algorithm*}

\bfsection{Cyclic graph} If the reduced graph has circles, then we assign a higher priority to the circles and name the graph accordingly. For the graph structure with one circle, the prefix is \textbf{\emph{cyclo}}. We then name branches use parentheses containing relative location on the ring together with the branch description described before. For a bicyclic graph where two circles share at least one vertexes, the root numeral prefix of the graph name depends on the total number of vertexes in all rings together\cite{favre2013nomenclature}. The prefix \textbf{\emph{bicyclo}} denote the sharing of at least two vertexes, while \textbf{\emph{spiro}} denote the sharing of only one vertex. In between the prefix and the suffix, a pair of brackets with numerals denotes the number of vertexes between each of the bridgehead ones. These numbers are arranged in descending order and are separated by periods. For example, a graph with a 3-vertex circle and a 5-vertex circle share two vertexes (one edge) will be named $bicyclo[3,1,0]hexito$ (Fig.~\ref{fig:method}). Such rules can be easily extrapolated into graphs with more than three circles. However, in practice, we notice rare cases in the mitochondria and pyramidal neurons with more than two circles.



\subsection{Topological Nomenclature Embedding}
In this subsection, we describe how to extract features from the reduced graphs to estimate the commonality/dissimilarity between them. To represent a graph as a matrix, we construct an adjacency matrix, whose elements are connectivities between nodes in a graph. We employ a variational graph autoencoder (VGAE)~\cite{kipf2016variational}, which is a neural network for unsupervised learning on graphs based on a variational autoencoder~\cite{kingma2013auto}, to extract features for each adjacency matrix.

We first normalize the adjacency matrix using the symmetric normalization scheme, as done in VGAE~\cite{kipf2016variational}. Then, we feed the normalized adjacency matrix into VGAE consisting of two graph convolutions and one fully connected layer to reconstruct that matrix at the output. The dimensions of the two graph convolutions are 32 and 16, respectively. We train the network for 200 epochs to minimize the difference between the input adjacency matrix and the reconstructed one and the Kullback-Leibler (KL) divergence of the embedding. We use Adam optimizer with a learning rate of 0.01. We exploit the embedding output of the graph convolutions as the nomenclature embedding.

\begin{figure*}[t]
     \centering
     \includegraphics[width=0.8\textwidth]{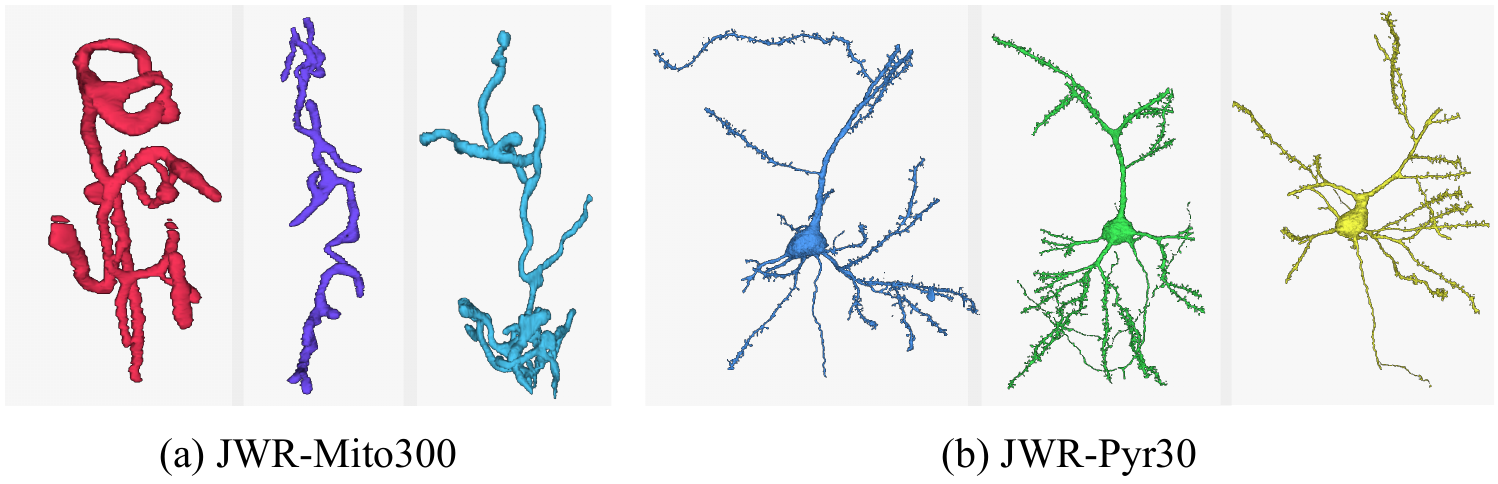}
     \caption{Samples in our JWR-Mito300 and JWR-Pyr30 datasets. (a) Unlike textbook illustrations, mitochondria can have complicated 3D shapes. (b) Pyramidal neurons exhibit great diversity in the spatial distribution of dendrites and their branching patterns.}
     \label{fig:dataset}
\end{figure*}

\section{Dataset}
By inspecting the shape of pyramidal cells and mitochondria at nanometer resolution, we found hundreds of structures that are very different from textbook illustrations, which usually display simple spherical or tubular structures. To further advance shape analysis in EM connectomics, we build a dataset of non-trivial objects to exhibit the complexity of neuronal structures and test our topological nomenclature system. We will release the dataset publicly.

\bfsection{Data Acquisition} 
We imaged a tissue block from Layer II/III in the primary visual cortex of an adult rat at a resolution of $4\times4\times30~nm^3$ using a multi-beam scanning electron microscope (EM). After stitching and aligning the 2D images using multi-CPU clusters, we obtained a final 3D image stack of 100 $\mu m$ cube.

\bfsection{3D Object Segmentation} 
Annotating the instances manually from scratch is not feasible. However, the accuracy of existing automatic segmentation algorithms can not generate object masks that are qualified enough for downstream morphological analysis. To have a good tradeoff between segmentation quality and efficiency, we first adopt the 3D U-Net model~\cite{ronneberger2015u} for initial automatic neuron and mitochondria segmentation.
We then use a manual annotation tool~\cite{berger2018vast} to proofread and modify the segmentation results.

\bfsection{JWR-Mito300}
We reconstructed all the mitochondria found in the somata of 11 cells: one pyramidal neuron, six interneurons, and four glial cells. 
Out of thousands of mitochondira,  we selected 316 of them that have nontrivial topological structures with a volumetric size larger than 0.2 $\mu m^{3}$ (Fig.~\ref{fig:dataset}\red{a}).

\bfsection{JWR-Pyr30}
For this dataset, we randomly selected 30 pyramidal cells whose cell bodies are located in the central volume with the presence of a significant portion (if not full) of their basal dendrites. Individual pyramidal neurons have one apical dendrite pointing to the pial surface and an axon often extending in the opposite direction. 
Nevertheless, they all show distinct distributions of oblique and basal dendrites (Fig.~\ref{fig:dataset}\red{b}).
\section{Experiments}\label{sec:exp}
In this section, we first quantitatively evaluate our nomenclature extraction results in terms of the agreement with human perception on our JWR-Mito300 and JWR-Pyr30 datasets. We then show qualitative results for two applications of the extracted nomenclature feature, including 3D shape retrieval and 3D shape decomposition. 

\begin{table}[t]
\centering
\caption{Ablation studies on graph reduction parameters on JWR-Mito300 and JWR-Pyr30 dataset. We compute the average cosine similarity of the graph Laplacian eigenvalues and the accuracy (correct if cosine similarity is bigger than 0.95) for different design choices.}
\label{tab:ablation}
\resizebox{1.0\columnwidth}{!}{
\begin{tabular}{llcccc}
\hline
\multirow{2}{*}{Dataset}&\multirow{2}{*}{Metric} & \multicolumn{2}{c}{BFS+Junction} & \multicolumn{2}{c}{BFS+Junction+Loop}\\\cline{3-6}
 && $\tau$=0& $\tau$=4& $\tau$=0& $\tau$=4\\
\hline
 \multirow{2}{*}{JWR-Mito300} & Cosine & 0.891 & 0.949 & 0.914 & \textbf{0.956}\\
 & Acc. & 0.376 & 0.785 & 0.493 & \textbf{0.854}\\
 \multirow{2}{*}{JWR-Pyr30} & Cosine & 0.778 & \textbf{0.965} & N/A & N/A\\
 & Acc. & 0 & \textbf{0.767} & N/A & N/A\\
\hline
\end{tabular}}
\vspace{-0.1in}
\end{table}

\subsection{Nomenclature Extraction}
\bfsection{Experimental Setup}
As there is no known previous work that has similar experiments, we conducted ablation studies on different design choices of our proposed method.
To generate the reduced graph, we started from the base method of the modified BFS, referred to as {\em BFS+Junction}, and compared two design choices. One is the 
edge length threshold $\tau$ for edge contraction and the other is to preserve loop structure or not.
Note that in the JWR-Pyr30 dataset, pyramidal neurons usually do not form loops and we only compare different $\tau$ values.

\begin{figure*}[t]
     \centering
     \includegraphics[width=1.0\textwidth]{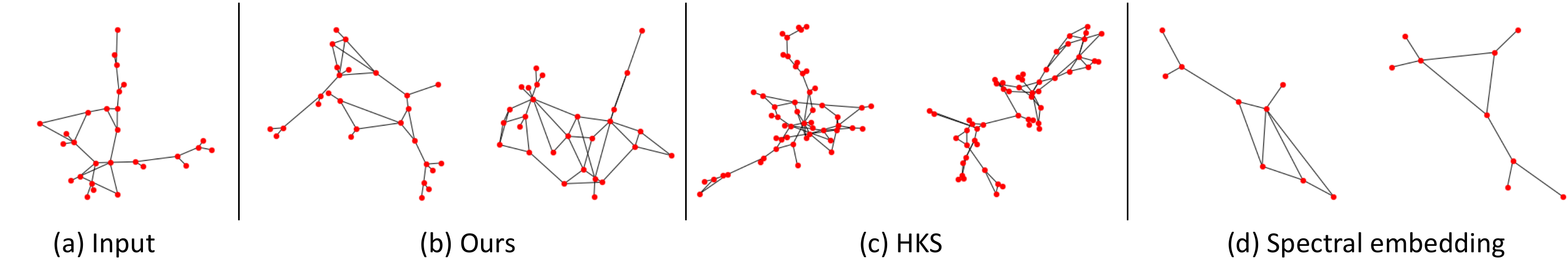}
     \caption{Shape retrieval using graph features. Given (a) an input graph representing the original 3D mesh, we show top-2 retrieval results using (b) our topological nomenclature approach, (c) Heat Kernel Signature (HKS), and (d) spectral embedding. The structures retrieved by our method are qualitatively more similar to the query sample.}
     \label{fig:retrieval}
\end{figure*}

\bfsection{Evaluation Metric}
To create ground truth nomenclature extraction labels for JWR-Mito30 and JWR-Pyr30 datasets,
we asked neuroscientists to draw their perceived planar graph representation when showing them with the original 3D object meshes.
To quantitatively evaluate our automatic nomenclature extractions, we use {\em Cosine}, the cosine similarity of the graph Laplacian eigenvalues between the prediction and the ground truth, as the metric.
Empirically, we found that human labeling results have around 0.95 cosine similarity due to the inherent ambiguity for small branches. Thus, we define a prediction accuracy metric, {\em Acc.}, with 0.95 cosine similarity as a threshold for correctness. 


\bfsection{Quantitative Results}
As shown in Table~\ref{tab:ablation}, the choice of threshold $\tau$ for small edge contractions is crucial, as the prediction accuracy almost doubles with $\tau=4.0$ compared to $\tau=0$. 
Although all design choices have similar average cosine similarity, they have a different number of correct predictions that are acceptable for downstream analysis.
Especially for pyramidal cells from the JWR-Pyr30 dataset, the post-processing parameter $\tau=4$ helps to remove many small spine structures.

Preservation of loops seems to have a small impact on overall performance, but it is still essential for
preserving the topology of cyclic mitochondria even if they do not appear often.
With the best design choices,  our automatic nomenclature extraction method can capture the essential topology of 3D complex shapes without producing disturbing artifacts at around 80\% accuracy.

Those results indicate that the graph extraction algorithm in the nomenclature system produces high-quality representations that are consistent with human perception. Considering that the nomenclature rule is designed to ensure that every name refers to a specific structure without ambiguity, those results further support the informativeness and conciseness of the assigned name for scientific communication. 
 


\subsection{Shape Retrieval}
\bfsection{Experimental Setup}
For a given query 3D shape, users may want to find similar shapes from the entire dataset. To this end, we perform 3D shape retrieval using our topological nomenclature. The goal of this experiment is to find two topological shapes that are most similar to a given query 3D shape. We use the JWR-Mito300 dataset and randomly sample a query 3D shape from that dataset. We discover two nearest neighbors of the query. 

To compare two 3D shapes, we first measure pairwise differences between their nomenclature embeddings by computing $L2$ distances. Then, we determine a similarity between them as an average of matching costs. We use Hungarian matching.

\bfsection{Qualitative Results}
Fig.~\ref{fig:retrieval} shows 3D shape retrieval results of our algorithm compared to both HKS~\cite{article} and spectral embedding~\cite{ng2002spectral}. The results exemplify that our algorithm is capable of discovering topologically similar 3D shapes. In contrast, HKS finds 3D shapes that have visually similar meshes but different actual neuronal or mitochondria structures. Since the spectral embedding encodes the entire graph, it fails to find relevant shapes.





\begin{figure*}[t]
     \centering
     \includegraphics[width=0.8\textwidth]{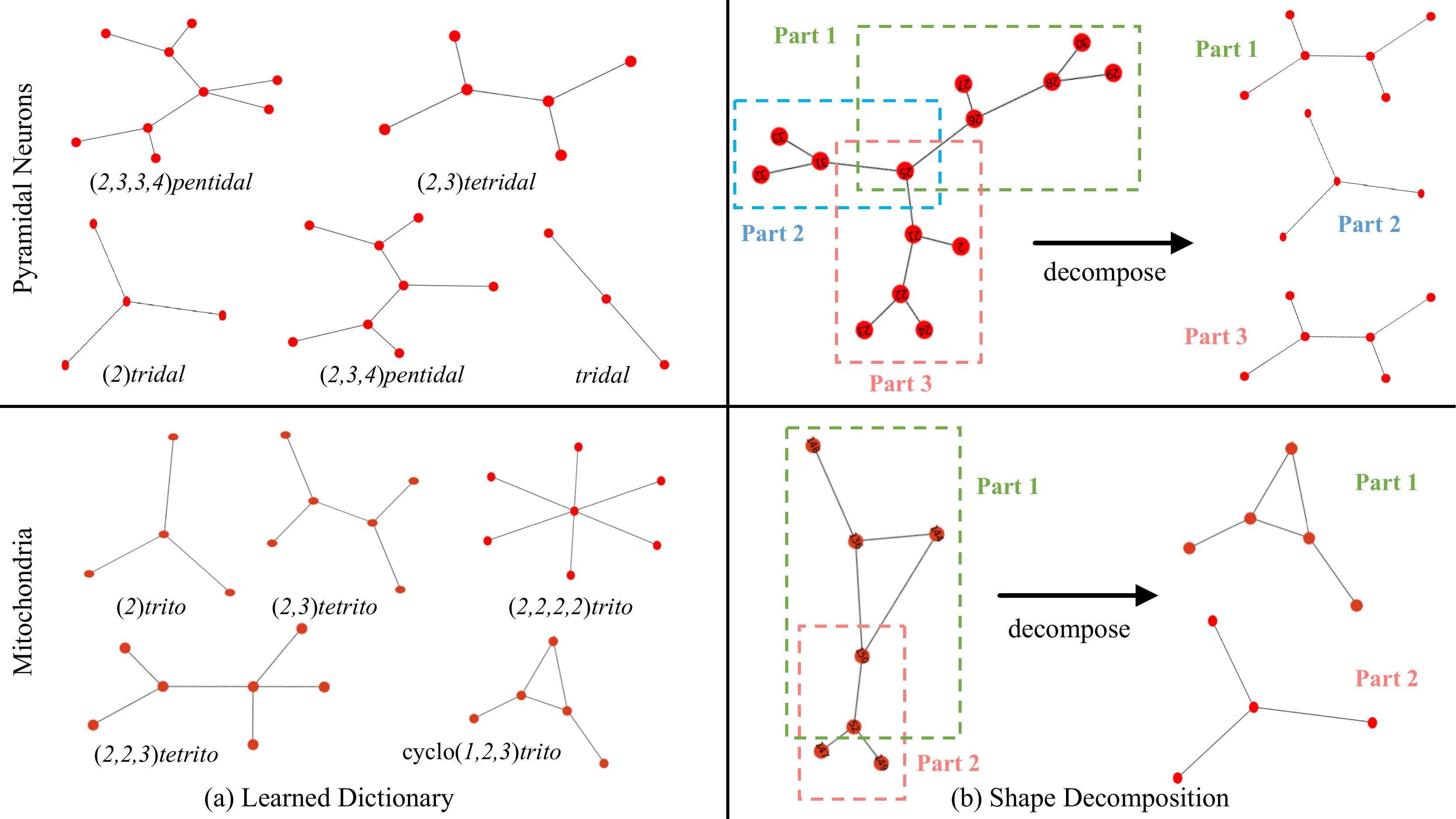}
     \caption{3D shape decomposition with topological nomenclatures. For mitochondrion (top row) and a pyramidal neurons (bottom row), we show (a) the learned part dictionary and (b) greedy decomposition result for an input example.}
     \label{fig:decomp}
\end{figure*}
\subsection{Shape Decomposition Results}
\bfsection{Experimental Setup}
Decomposing topological nomenclatures into sub-graphs enables users to understand the structures of 3D shapes. To define sub-graphs, we construct a dictionary of our nomenclature embedding features. Specifically, we apply $k$-means clustering algorithm to embedding features of \textit{junctions} (defined in Section~\ref{subsec:graphgen}) to generate words in the dictionary. We set $k$ as 50 and 100 for the pyramidal neurons and mitochondria, respectively. Note that we only use the junctions since end nodes have no local structures. In the inference phase of decomposition, we perform matching between junctions in a query nomenclature and the words in the dictionary. We first find a junction with the minimum distance, and then remove it and its neighbor nodes from the query nomenclature. We iterate this process until there are no more junctions.

\bfsection{Qualitative Results}
Fig.~\ref{fig:decomp}\red{a} visualizes dictionaries learned on the JWR-Mito300 and JWR-Pyr30 datasets. It is observable that the words in the dictionaries vary. Fig.~\ref{fig:decomp}\red{b} shows decomposition results of our nomenclatures. Our method precisely decomposes the nomenclatures into sub-graphs. 



\section{Conclusion}
In this paper, we introduced the topological nomenclature protocol to extract, name, and manipulate the morphology of biological objects in EM connectomics. We demonstrated the effectiveness of our novel nomenclature system through quantitative ablation studies. Moreover, our unsupervised nomenclature embedding successfully performed retrieval and decomposition of 3D shapes. We will make the two datasets containing 316 mitochondria with complex morphology and 30 pyramidal neurons publicly available. For future work, we will apply our nomenclature scheme to a large-scale dataset to understand the diversity and similarity of biological structures.

{\small
\bibliographystyle{ieee_fullname}

}

\end{document}